\documentclass{article}

\usepackage[preprint]{neurips_2026}

\usepackage[utf8]{inputenc} 
\usepackage[T1]{fontenc}    
\usepackage{hyperref}       
\usepackage{url}            
\usepackage{booktabs}       
\usepackage{amsfonts}       
\usepackage{nicefrac}       
\usepackage{microtype}      
\usepackage{xcolor}         

\usepackage{amsmath,amsfonts,bm}

\def\eqref#1{equation~\ref{#1}}

\def\1{\bm{1}}

\DeclareMathAlphabet{\mathsfit}{\encodingdefault}{\sfdefault}{m}{sl}
\SetMathAlphabet{\mathsfit}{bold}{\encodingdefault}{\sfdefault}{bx}{n}

\usepackage{hyperref}
\usepackage{url}
\usepackage{times}
\usepackage{amsmath, amssymb, amsfonts}
\usepackage{booktabs}
\usepackage{multirow}
\usepackage{graphicx}
\usepackage{xcolor}
\usepackage{url}
\usepackage{hyperref}
\usepackage{enumitem}
\usepackage{listings}
\usepackage{wrapfig}

\usepackage{tcolorbox}
\usepackage{xcolor}
\usepackage{enumitem}
\usepackage{soul}

\hypersetup{
    colorlinks=true,
    linkcolor=blue,
    citecolor=blue,
    urlcolor=blue
}

\title{TEFM: Token-Efficient Faithful 
Modeling \\ for 
Structured Data}

\author{
  Zhichao Hou\textsuperscript{\rm 1}
  \And
  Lingdao Sha\textsuperscript{\rm 2} 
  \And
  Xueyu Mao\textsuperscript{\rm 2}
  \And
  Yang Liu\textsuperscript{\rm 2}
  \And
  Peijie Qiu\textsuperscript{\rm 2}
  \And
  Rui Song\textsuperscript{\rm 2}\thanks{Corresponding author.}  
  \\
    \and
  \textsuperscript{\rm 1}North Carolina State University, 
  \textsuperscript{\rm 2}Amazon Web Services\\
  \and
  \small{\texttt{zhou4@ncsu.edu, 
  \{lingdao,maxueyu,yngliun,peijieq,ruisong\}@amazon.com}}\\
}

\begin{document}

\maketitle

\begin{abstract}
In this paper, we solve two fundamental obstacles in applying LLMs to critical domains: 
token efficiency and faithfulness. To address both constraints jointly, we present 
\textbf{TEFM} (Token-Efficient Faithful Modeling), a framework designed for structured 
data analysis in critical domains. TEFM achieves token efficiency by compressing lengthy 
structured observations into compact Behavioral Code tokens, dramatically reducing token 
consumption with minimal information loss. Moreover, TEFM enables faithful rationalization 
through a dual-fidelity objective that jointly optimizes code-level reconstruction and 
prediction-level fidelity, identifying minimal sufficient feature subsets grounded in input 
data. Comprehensive experiments across various domain datasets and 
model backbones (Qwen3, Gemma-2, Phi-4) show that TEFM achieves competitive classification 
accuracy with dramatic token reduction (approximately 1\% token retention in clinical and 2\% 
in security domains) while producing faithful rationales.
\end{abstract}
\section{Introduction}
\label{sec:introduction}

Large language models have demonstrated remarkable capability across diverse critical domains. 
In healthcare, LLMs can assist diagnosis by analyzing lengthy medical records containing lab values, 
vital signs, and historical diagnoses~\citep{rajkomar2018scalable}. In cybersecurity, LLMs can detect 
intrusions by examining network flow characteristics~\citep{zarpelao2017survey}. In marketplace regulation, 
LLMs can enforce policies by reviewing transaction histories.

\begin{wrapfigure}{r}{0.4\textwidth}
    \centering
    \vspace{-0.15in}
    \includegraphics[width=0.4\textwidth]{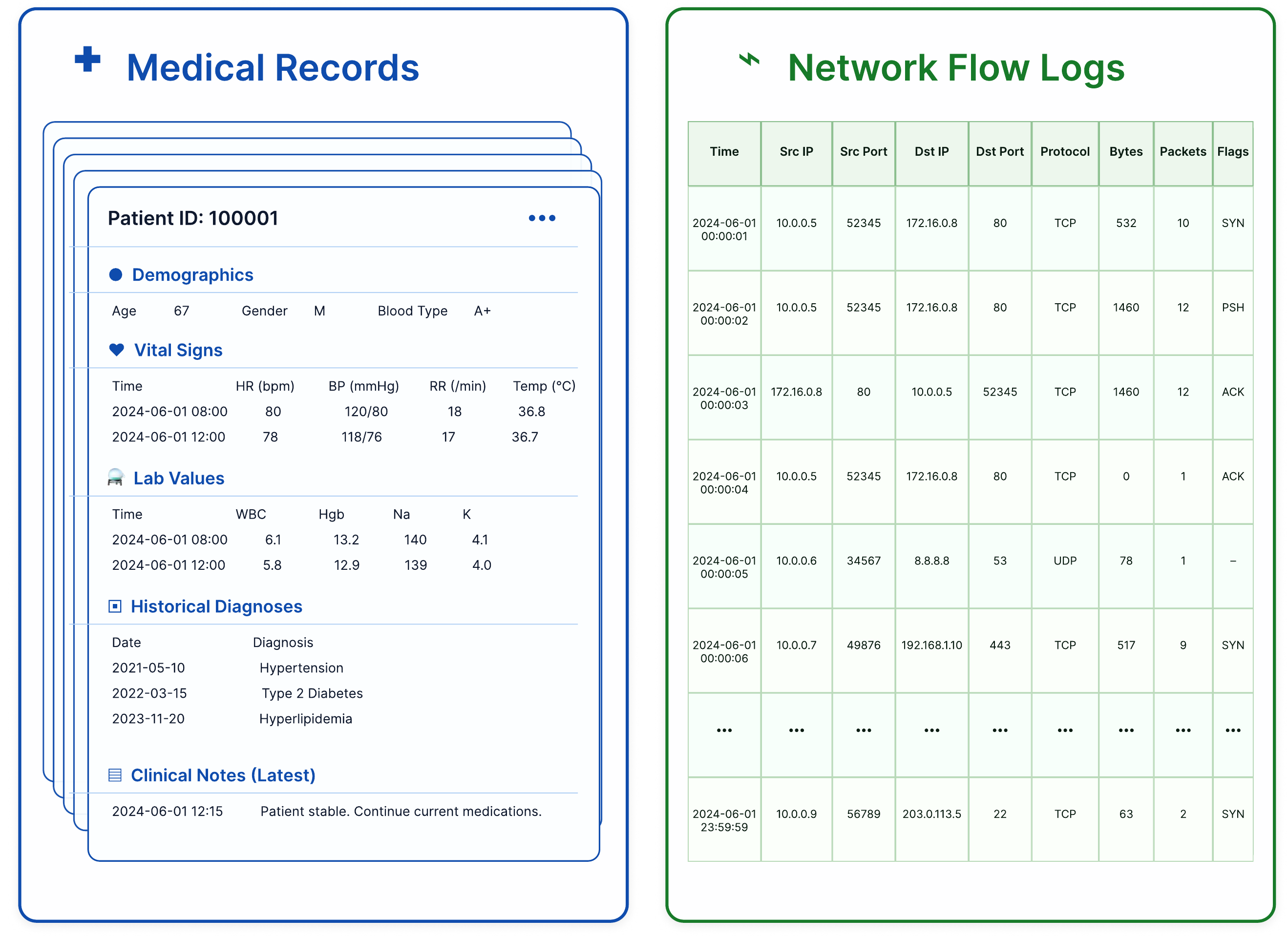}
    \caption{Structured data examples.}
    \label{fig:structured_data}
    \vspace{-0.2in}
\end{wrapfigure}

As shown in Figure~\ref{fig:structured_data}, these domains typically generate structured data 
such as medical records with patient vitals and lab values, or network flow logs with packet-level 
statistics. While conventional models (tree-based ensembles, feedforward networks) handle tabular 
data effectively, LLM-based approaches offer complementary advantages in safety-critical applications: 
(1)~\emph{multimodal reasoning}---combining structured data with clinical notes, incident logs, or other 
text without separate preprocessing; (2)~\emph{in-context interpretability}---LLMs naturally produce 
human-readable reasoning grounded in specific features; (3)~\emph{unified infrastructure}---organizations 
already deploying LLMs can leverage a single model across domains. Enabling LLMs to reason over structured 
data therefore opens a promising direction for safety-critical prediction.

However, applying LLMs to structured data introduces two fundamental obstacles:

\noindent\textbf{(1) Token Efficiency.}
Structured data such as medical records or network logs consume excessive tokens when serialized as text. 
Long observation sequences quickly exhaust available context windows. The proportional cost of API calls 
and memory overhead make it impractical to process lengthy sequences without careful management of token consumption.

\noindent\textbf{(2) Faithfulness.}
LLMs are prone to hallucinations and unfaithful predictions. In safety-critical domains such as healthcare 
and security, unreliable outputs pose severe risks: incorrect diagnoses may delay treatment, and false alerts 
may trigger costly interventions. Beyond accuracy alone, users require evidence that predictions are grounded 
in input data to support trustworthy decision-making.

Prior work has addressed these challenges separately. Context-extension approaches such as global memory 
tokens~\citep{beltagy2020longformer,ainslie2020etc,zaheer2020big}, sparse attention mechanisms~\citep{child2019generating,roy2021efficient,kitaev2020reformer}, 
and segment-based recurrence~\citep{dai2019transformer,rae2019compressive} modify model architecture but still 
scale with sequence length and suffer information loss under aggressive compression. Rationalization methods 
aim to identify sufficient input features~\citep{lei2016rationalizing,bastings2019interpretable,camburu2018snli,narang2020wt5}, 
but most focus on NLP tasks with short free-text inputs; faithful rationalization for structured data remains 
largely unexplored.

This raises a fundamental question: \emph{How can we build a model that is simultaneously token-efficient, 
predictively accurate for high-dimensional structured data, and capable of providing faithful rationales 
grounded in the input?}

To this end, we propose \textbf{TEFM} (Token-Efficient Faithful Modeling), a framework for structured data. 
TEFM reduces token consumption at the source with minimal information loss while providing faithful 
rationales grounded in input data. Our contributions are:

\begin{itemize}[leftmargin=*]
    \item A hierarchical tokenization pipeline that compresses lengthy structured observations into compact 
    behavioral codes via Residual Quantized VAE, achieving $\sim$99\% token reduction with minimal information loss.

    \item A two-stage training recipe: BC-text alignment pretraining integrates behavioral codes with LLM 
    embeddings, preserving high-fidelity representations; supervised fine-tuning trains the LLM on compressed 
    codes as a replacement for raw input, enabling token-efficient accurate predictions.

    \item A dual-fidelity rationalization framework that identifies minimal sufficient feature subsets by jointly 
    optimizing code-level reconstruction and prediction-level fidelity, yielding faithful explanations grounded 
    in input data without post-hoc perturbation.

    \item Comprehensive evaluation on clinical mortality prediction (MIMIC-III) and network intrusion detection 
    (CIC-IDS2017), demonstrating competitive accuracy across diverse LLM architectures (Qwen3, Gemma-2, Phi-4), 
    robust cross-domain generalization, and faithful rationale extraction.
\end{itemize}

The remainder of this paper is organized as follows. Section~\ref{sec:method} describes the TEFM architecture, 
training methodology, and rationalization framework. Section~\ref{sec:experiments} presents experimental results 
on clinical and security domains with thorough ablations. Section~\ref{sec:conclusion} discusses limitations and 
future directions.

\begin{figure}[h!]
    \centering
    \includegraphics[width=0.99\textwidth]{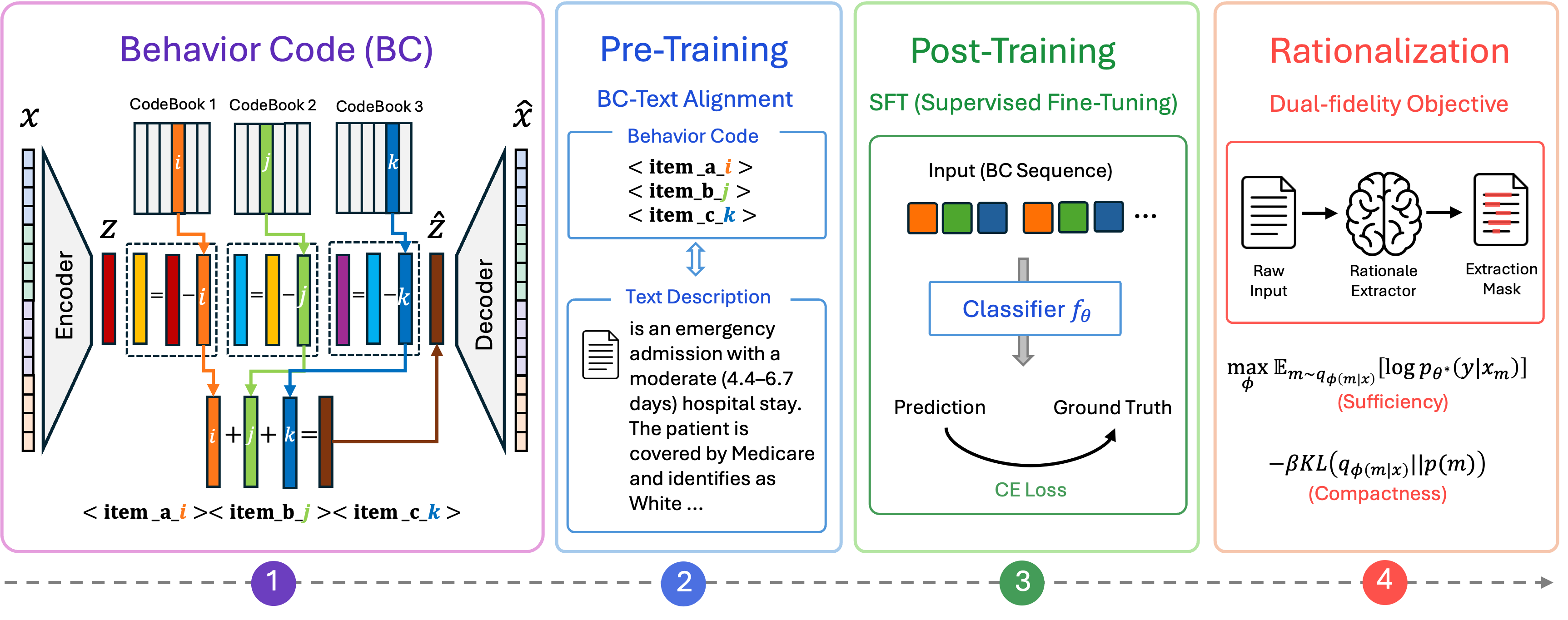}
    \caption{TEFM pipeline. Structured observations are compressed into discrete hierarchical Behavioral Codes (BCs) by an RQ-VAE tokenizer, dramatically reducing context cost. BC tokens are aligned with an LLM via lightweight pretraining, enabling efficient prediction over observation sequences. A learned feature extractor then provides interpretable rationales grounded in the original input, without requiring additional LLM inference.}
    \label{fig:pipeline}
\end{figure}

\section{TEFM: Token-Efficient Faithful Modeling}
\label{sec:method}

In this section, we introduce TEFM, a unified framework comprising four stages:
(1) hierarchical Behavioral Code (BC) tokenization,
(2) BC--text alignment pretraining,
(3) supervised fine-tuning for classification, and
(4) dual-fidelity variational rationalization.
Figure~\ref{fig:pipeline} provides an overview of the complete pipeline.

\subsection{Hierarchical High-Fidelity Tokenization}
\label{sec:tokenization}


In many critical domains, structured data—such as medical records and network flow telemetry—are ubiquitous. A naive approach would serialize all observations into natural language and feed them into an LLM. However, this is prohibitively expensive: high-dimensional records quickly exhaust token budgets and incur substantial computational costs.
Fortunately, structured observations exhibit inherent structure we can exploit. In reality, schema-conforming records exhibit natural hierarchy: groups of features co-occur, and many observations share dominant patterns while differing along only a few dimensions. Rather than treating each observation independently, we can leverage this hierarchical structure to dramatically reduce redundancy.
We therefore propose hierarchical behavioral tokenization to exploit this structure and compress lengthy observations into a small set of hierarchical codes. This approach dramatically reduces token consumption while preserving task-relevant information for accurate predictions.

\begin{wrapfigure}{r}{0.45\textwidth}
\vspace{-0.25in}
\centering
\[
\begin{array}{lll}
\mathbf{x}_1 = [0,0,0,0,0,0] & \!\rightarrow\! & \langle a_0\rangle\langle b_0\rangle\langle c_0\rangle \\
\mathbf{x}_2 = [0,0,0,0,1,1] & \!\rightarrow\! & \langle a_0\rangle\langle b_0\rangle\langle c_1\rangle \\
\mathbf{x}_3 = [0,0,1,1,0,0] & \!\rightarrow\! & \langle a_0\rangle\langle b_1\rangle\langle c_0\rangle \\
\mathbf{x}_4 = [0,0,1,1,1,1] & \!\rightarrow\! & \langle a_0\rangle\langle b_1\rangle\langle c_1\rangle \\
\mathbf{x}_5 = [1,1,0,0,0,0] & \!\rightarrow\! & \langle a_1\rangle\langle b_0\rangle\langle c_0\rangle \\
\mathbf{x}_6 = [1,1,0,0,1,1] & \!\rightarrow\! & \langle a_1\rangle\langle b_0\rangle\langle c_1\rangle \\
\mathbf{x}_7 = [1,1,1,1,0,0] & \!\rightarrow\! & \langle a_1\rangle\langle b_1\rangle\langle c_0\rangle \\
\mathbf{x}_8 = [1,1,1,1,1,1] & \!\rightarrow\! & \langle a_1\rangle\langle b_1\rangle\langle c_1\rangle \\
\end{array}
\]
\vspace{-0.2in}
\caption{Eight length-6 binary observations are compressed into 3-level hierarchical codes.}
\label{fig:toy}
\vspace{-0.15in}
\end{wrapfigure}

\textbf{Toy Example.} To better illustrate this hierarchical compression principle, we present a concrete toy example in Figure~\ref{fig:toy}.
Consider eight binary observations of length 6. The first two dimensions always agree, so one top-level code $\langle a \rangle$ suffices: $\{\mathbf{x}_1,\ldots,\mathbf{x}_4\}$ share $\langle a_0\rangle$ and $\{\mathbf{x}_5,\ldots,\mathbf{x}_8\}$ share $\langle a_1\rangle$. Conditioned on $\langle a \rangle$, the next two dimensions co-occur and are captured by a second-level code $\langle b \rangle$; the final two dimensions by $\langle c \rangle$. With 3 layers of codebook size 2, only $3 \times 2 = 6$ vocabulary tokens encode all $2^3 = 8$ patterns, with each observation compressed to exactly 3 tokens.
In practice, the input space is vast in dimension and the hierarchy far more complex. However, the principle generalizes: behavioral data exhibits coarse-to-fine structure, and models that discover this structure automatically can represent exponentially large pattern spaces with tiny, reusable vocabularies.
In general, a $K$-layer codebook with $C$ entries per layer requires only $K \times C$ vocabulary tokens yet distinguishes up to $C^K$ distinct patterns. With $K=3$ and $C=256$, only 768 tokens encode up to $256^3 \approx 16.7$M patterns, and each observation can be represented by exactly $K$ tokens.

\textbf{Hierarchical High-Fidelity Tokenization.}
Despite this intuition, achieving high-fidelity compression with efficient codebook utilization is non-trivial. Our primary goal is to dramatically reduce token consumption while preserving the information necessary for accurate predictions. The encoding process must not discard critical details and we should be able to recover the essential information from the compressed tokens. While prior work has attempt to create discrete codes for image~\citep{van2017neural} and descriptive sentence~\citep{hou2023learning}, those methods focus on general descriptive information without stringent requirements for lossless recovery or minimal information loss. In contrast, our setting demands faithful reconstruction: the compressed codes must retain sufficient information to support high-fidelity predictions from limited tokens.

To achieve high-fidelity hierarchical tokenization, we first vectorize structured records: categorical fields are enumerated, and continuous fields are discretized into $B$ equal-frequency buckets, without preserving field names or metadata since this information is shared across all  records and can be memorized by LLM during pretraining.
Then the resulting vector $\mathbf{x} \in \mathbb{R}^D$ is encoded via $\mathbf{E}(\cdot): \mathbb{R}^D \to \mathbb{R}^d$  and reconstructed via $\mathbf{D}(\cdot): \mathbb{R}^d \to \mathbb{R}^D$ via Residual Quantized VAE (RQ-VAE) architecture as shown in Figure~\ref{fig:pipeline} (Step 1).
Given $\mathbf{e} = \mathbf{E}(\mathbf{x})$, we quantize through $K$ residual levels. At level $k$, the residual $\mathbf{r}^{(k)}$ (with $\mathbf{r}^{(1)} = \mathbf{e}$) is assigned to its nearest codebook entry:
\begingroup
\setlength{\abovedisplayskip}{5pt}
\setlength{\belowdisplayskip}{5pt}
\setlength{\abovedisplayshortskip}{5pt}
\setlength{\belowdisplayshortskip}{5pt}
\[
i_k = \arg\min_j
\left\lVert \mathbf{r}^{(k)}-\mathbf{e}_j^{(k)} \right\rVert_2^2,
\qquad
\hat{\mathbf{r}}^{(k)}=\mathbf{e}_{i_k}^{(k)},
\qquad
\mathbf{r}^{(k+1)}
=\mathbf{r}^{(k)}-\hat{\mathbf{r}}^{(k)}.
\]
\endgroup
The quantized embedding $\hat{\mathbf{e}} = \sum_{k=1}^{K} \hat{\mathbf{r}}^{(k)}$ is decoded to reconstruct the observation. The residual structure realizes the hierarchy: level 1 captures dominant patterns, and each subsequent level refines it.
Training optimizes reconstruction plus per-level codebook and commitment losses:
\begingroup
\setlength{\abovedisplayskip}{5pt}
\setlength{\belowdisplayskip}{5pt}
\setlength{\abovedisplayshortskip}{5pt}
\setlength{\belowdisplayshortskip}{5pt}
\[
\mathcal{L}
=
\left\| \mathbf{D}(\hat{\mathbf{e}}) - \mathbf{x} \right\|_2^2
+
\sum_{k=1}^{K}
\left[
\left\| \hat{\mathbf{r}}^{(k)} - \operatorname{sg}[\mathbf{r}^{(k)}] \right\|_2^2
+
\beta \left\| \mathbf{r}^{(k)} - \operatorname{sg}[\hat{\mathbf{r}}^{(k)}] \right\|_2^2
\right],
\]
\endgroup
where $\operatorname{sg}[\cdot]$ is the stop-gradient operator and $\beta$ weights the commitment loss. Through vectorization and an encoder-decoder architecture, we compress each lengthy record into $K$ discrete behavioral codes (BC) $\langle i_1 \rangle, \langle i_2 \rangle, \ldots, \langle i_K \rangle$ using encoder $\mathbf{E}$, then recover the original data via decoder $\mathbf{D}$.
\vspace{-0.15in}
\subsection{Token-Efficient Training}
After obtaining the behavioral codes (BC), we need to perform token-efficient prediction without directly accessing the original lengthy records. To achieve this, we first perform BC-text alignment to help the LLM understand these special tokens. Then we perform supervised fine-tuning on the prediction task of interest.

\textbf{Pre-training: BC-Text Alignment.}
Behavioral Codes are new tokens outside the base LLM's vocabulary. We align them through lightweight pretraining on paired (BC, natural-language description) data.
We instantiate TEFM on a base LLM (e.g., Qwen3-1.7B) and add $K \times C + 2$ new tokens: the $K \times C$ BC tokens plus boundary markers. Only the embedding layer is trainable; all transformer layers remain frozen. Since only $\sim$1--2\% of parameters are trainable, pretraining converges in a small number of epochs.
Each training example pairs a behavioral code with a natural-language description:
\begin{small}
\vspace{-0.3in}
\begin{tcolorbox}[colback=white, 
                   colframe=black,
                   coltitle=white,
                   colbacktitle=black!75,
                   title=Pre-training Example in Clinical Data,
                   boxrule=1.5pt,
                   width=1.0\textwidth,
                   fonttitle=\bfseries,
                   arc=0pt,
                   top=8pt,
                   bottom=8pt,
                   left=10pt,
                   right=10pt]
\footnotesize
\begin{verbatim}
This admission <|item_begin|><adm_a_138><adm_b_231><adm_c_132><|item_end|> is 
an emergency admission with a moderate (4.4-6.7 days) hospital stay. The 
patient is covered by Medicare and identifies as White...
\end{verbatim}

\end{tcolorbox}
\vspace{-0.1in}
\end{small}

\textbf{Fine-tuning: Task-specific SFT.} After pretraining, we fine-tune the LLM on downstream classification tasks using BC-encoded representations as input. The model performs binary or multi-class prediction over observation sequences. For mortality prediction, the input presents the patient's complete admission history and asks whether the patient died during their final hospitalization. For intrusion detection, the input presents a sequence of network flows and asks whether the final flow is part of an attack. Below we illustrate an SFT example from the clinical domain:

\begin{small}
\vspace{-0.01in}
\begin{tcolorbox}[colback=white, 
                   colframe=black,
                   coltitle=white,
                   colbacktitle=black!75,
                   title=SFT Example in Clinical Data,
                   boxrule=1.5pt,
                   width=\textwidth,
                   fonttitle=\bfseries,
                   arc=0pt,
                   top=8pt,
                   bottom=8pt,
                   left=10pt,
                   right=10pt]

\footnotesize
\texttt{\begin{tabbing}
\textbf{<|im\_start|>user} \\
This patient has 3 hospital admission(s):<|item\_begin|><adm\_a\_68><adm\_b\_25>\\
<adm\_c\_7><|item\_end|>, <|item\_begin|><adm\_a\_12><adm\_b\_91><adm\_c\_44><|item\_end|>,\\
<|item\_begin|><adm\_a\_53><adm\_b\_67><adm\_c\_19><|item\_end|>. Based on the\\ 
patient's admission history, did the patient die during their 
last hospital \\
admission? \\
\textbf{<|im\_end|>} \\
\textbf{<|im\_start|>assistant} \\
Yes, the patient died. \\
\textbf{<|im\_end|>}
\end{tabbing}}

\end{tcolorbox}
\end{small}

\subsection{Dual-Fidelity Rationalization}
In critical domains, predictions must be grounded in transparent rationales: 
users require evidence that decisions are justified by the input data. 
We formalize this requirement through two design principles:
(1) Sufficiency: The extracted rationale must preserve predictive power—removing 
the selected features should not significantly degrade model performance.
(2) Compactness: The rationale should be minimal—identifying only the essential 
features necessary for prediction, not extraneous ones.
To achieve both principles simultaneously, we formalize rationalization as learning 
a binary mask that identifies the minimal input subset sufficient to preserve predictions. 
This mechanism forces the model to select only features that genuinely contribute to the decision.

\textbf{Variational Objective.}
Let $\mathbf{x} \in \mathbb{R}^D$ be the raw input, $\mathbf{c} = \operatorname{BC}(\mathbf{x}) \in \{0, \ldots, C-1\}^K$ its behavioral code, and $y$ the label. We seek a mask $\mathbf{m} \in \{0,1\}^D$ identifying features responsible for both code assignment and prediction:
\begin{align}
\max_{\phi} \quad &\mathbb{E}_{\mathbf{m} \sim q_\phi(\mathbf{m} \mid \mathbf{x})} \left[ \log p_{\theta^*}\left(y \mid \mathbf{x}_{\mathbf{m}}\right) \right] - \beta \operatorname{KL}\left(q_\phi(\mathbf{m} \mid \mathbf{x}) \,\|\, p(\mathbf{m})\right),
\end{align}
where $q_\phi(\mathbf{m} \mid \mathbf{x})$ is a learned rationale extractor, 
$p_{\theta^*}(y \mid \mathbf{x}_{\mathbf{m}})$ is the prediction model on masked inputs, 
and $p(\mathbf{m})$ is a sparse prior that encourages feature compactness. The first term ensures \emph{sufficiency}: the selected features must preserve predictive fidelity. The second term ensures \emph{compactness}: the mask is regularized toward sparsity, selecting only essential features.
However, this objective cannot be optimized directly: each candidate mask incurs 
the cost of RQ-VAE re-encoding and LLM inference, and both mask sampling and BC encoding 
are discrete operations that preclude gradient-based optimization.

\textbf{Tractable Dual-fidelity Objective.}
We exploit the conditional independence structure: since the LLM observes only $\mathbf{c}$, 
not $\mathbf{x}$ directly, we have $Y \perp X \mid C$. This yields:
\begin{equation}
p_{\theta^*}(y \mid \mathbf{x}) = \sum_{\mathbf{c}} p_{\theta^*}(y \mid \mathbf{c}, \mathbf{x})\, p_{\theta^*}(\mathbf{c} \mid \mathbf{x}) 
= \sum_{\mathbf{c}} p_{\theta^*}(y \mid \mathbf{c})\, p_{\theta^*}(\mathbf{c} \mid \mathbf{x}).
\end{equation}
Leveraging this decomposition, we can split the prediction fidelity into two complementary terms: 
(i) $p_{\theta^*}(\mathbf{c} \mid \mathbf{x})$ ensures the selected input preserves the original 
behavioral code; and (ii) $p_{\theta^*}(y \mid \mathbf{c})$ ensures consistent predictions on the 
compressed representation. 
In practice, we train an extractor network $f_\phi: \mathbb{R}^D \to [0,1]^D$ outputting a soft mask $\mathbf{z} = f_\phi(\mathbf{x})$. The masked input is $\mathbf{x}_{\mathbf{z}} = \mathbf{x} \odot \mathbf{z}$. The extractor is a residual MLP:$f_\phi(\mathbf{x}) = \sigma\left(\text{MLP}(\mathbf{x}) + W_{\text{skip}}\mathbf{x}\right),$
where $\text{MLP}$ is a three-layer network and $\sigma$ is sigmoid function. 
The overall loss combines three terms:
\begin{equation}
\mathcal{L}(\phi) = \lambda_1 \mathcal{L}_\text{BC} + \lambda_2 \mathcal{L}_\text{pred} + \lambda_3 \mathcal{L}_\text{sparse}.
\end{equation}

 In practice, we decompose the overall objective into three complementary terms: (1) BC Reconstruction Sufficiency: Ensures masked features preserve RQ-VAE latent representation:
$\mathcal{L}_\text{BC} = \left\| \mathbf{E}(\mathbf{x} \odot \mathbf{z}) - \operatorname{sg}[\mathbf{E}(\mathbf{x})] \right\|_2^2.$
(2) Prediction Sufficiency: Since the fine-tuned LLM is non-differentiable during extraction, we train a lightweight surrogate MLP classifier $g_\psi:$  on RQ-VAE latents to mimic LLM predictions. We minimize KL divergence between surrogate outputs on masked versus original inputs:
$\mathcal{L}_\text{pred} = D_\text{KL}\left(g_\psi(\mathbf{E}(\mathbf{x})) \,\big\|\, g_\psi(\mathbf{E}(\mathbf{x} \odot \mathbf{z}))\right).$
(3) Sparsity: Penalizes total mask mass to encourage compact explanations:
$\mathcal{L}_\text{sparse} = \|\mathbf{z}\|_1$.
Unlike the ideal variational objective, this formulation is fully differentiable, requires no stochastic sampling, and trains stably via backpropagation.

\textbf{Input-adaptive top-K masking.}
After training, we extract binary rationales via input-adaptive top-K masking. For each input, we compute the soft mask $\mathbf{z} = f_\phi(\mathbf{x})$ and select the top-$K$ features:
$m_i = \mathbf{1}[z_i \in \text{top-K}(\mathbf{z})],$
where $K$ is set to achieve a target sparsity level (e.g., 5--30\%). This allows each sample to use different features based on learned importance. The masked observations are re-encoded through the RQ-VAE to produce rationale-based Behavioral Codes, enabling evaluation of retained information at each sparsity level.
\section{Experiments}
\label{sec:experiments}
In this section, we evaluate TEFM from four complementary perspectives: (1) tokenization reconstruction fidelity, (2) token-efficient prediction performance on behavior codes, (3) rationalization quality, and (4) ablation studies that isolate the contribution of each training stage and model component.

\subsection{Experimental Settings}

\textbf{Datasets and Tasks.}
We evaluate TEFM on two domains: clinical mortality prediction and network intrusion detection. For mortality prediction, we use MIMIC-III~\citep{johnson2016mimic}, where each patient is represented by a sequence of hospital admissions with structured clinical records. For intrusion detection, we use CIC-IDS2017~\citep{sharafaldin2018toward}, focusing on DDoS and PortScan traffic represented by sequences of network flows.
Detailed dataset descriptions are provided in Appendix~\ref{sec:data_app}.

\textbf{Backbone and Model Configuration.}
Unless otherwise specified, we use Qwen3-1.7B as the default backbone. To evaluate the scalability and generality of TEFM, we further conduct experiments on Qwen3-0.6B, Qwen3-4B, Qwen3-8B, Gemma2-2B, and Phi-4-mini. All backbones use the same behavioral code tokenizer and training pipeline.

\textbf{Hyperparameters and Training.}
BC--text alignment pretraining is conducted for 3 epochs using AdamW with a learning rate of $10^{-3}$ and linear warmup over the first 500 optimization steps, while the transformer layers remain frozen. Supervised fine-tuning is conducted for 5 epochs using a causal language modeling objective, a learning rate of $10^{-5}$, a cosine annealing schedule, and full-parameter optimization. The rationalization extractor is trained for 100 epochs using Adam with a learning rate of $10^{-3}$. We set the fidelity-loss weights to $\lambda_1=\lambda_2=1.0$ and the sparsity-loss weight to $\lambda_3=0.1$. All experiments are conducted with a batch size of 32 for the clinical task and 64 for the network security tasks.

\subsection{Tokenization}

We evaluate the tokenization pipeline across three dimensions: (1) RQ-VAE compression fidelity, measuring reconstruction fidelity via slot accuracy; (2) BC-text alignment pretraining, verifying that the frozen encoder recovers codes from natural-language descriptions; and (3) hierarchical structure, confirming that multi-level organization captures meaningful domain similarity at progressively finer granularities. All evaluations use $K=3$ levels with $C=128$ codebook entries per level (K3-C128) as the default configuration; ablations across alternative codebook depths and sizes are in Section~\ref{sec:ablation}.

\begin{table*}[h!]
\centering
\begin{minipage}[t]{0.56\textwidth}
\centering
\caption{RQ-VAE reconstruction quality.}
\label{tab:rq-vae-default}
\small
\setlength{\tabcolsep}{4pt}
\begin{tabular}{lccc}
\toprule
\textbf{Dataset}
& \textbf{Slot AUC}
& \textbf{Slot AUC ($\pm 1$)}
& \textbf{Reconstruction} \\
\midrule
Clinical & 0.9438 & 0.9491 & 0.0407 \\
DDoS     & 0.9728 & 0.9994 & 0.0012 \\
PortScan & 0.9859 & 0.9995 & 0.0009 \\
\bottomrule
\end{tabular}
\end{minipage}
\hfill
\begin{minipage}[t]{0.40\textwidth}
\centering
\caption{BC-text alignment quality.}
\label{tab:align-default}
\small
\setlength{\tabcolsep}{4pt}
\begin{tabular}{lcc}
\toprule
\textbf{Domain}
& \textbf{Slot AUC}
& \textbf{Slot AUC ($\pm 1$)} \\
\midrule
Clinical & 0.9063 & 0.9177 \\
DDoS     & 0.9692 & 0.9976 \\
PortScan & 0.9845 & 0.9984 \\
\bottomrule
\end{tabular}
\end{minipage}

\end{table*}


\textbf{RQ-VAE Compression Fidelity.}
We use three metrics to measure compression fidelity: (1) Slot AUC measures the match 
ratio between original and reconstructed features; (2) Slot AUC ($\pm 1$) tolerates 
single-bucket quantization errors; (3) Reconstruction MSE quantifies normalized 
feature-level error. Table~\ref{tab:rq-vae-default} demonstrates high-fidelity 
reconstruction across domains.


\textbf{BC-Text Alignment Pretraining.}
BC tokens achieve strong alignment with the LLM embedding space through lightweight 
pretraining (Table~\ref{tab:align-default}). Clinical 
alignment reaches 90.6\%, security tasks exceed 96.9\%, with tolerant accuracy (Slot AUC $\pm 1$) 
exceeding 91.7\% across domains, confirming that the frozen encoder reliably maps textual 
descriptions to corresponding BCs.

\begin{wraptable}{r}{0.55\textwidth}
\centering
\vspace{-0.25in}
\caption{Hierarchical structure: intra-group distances decrease with longer BC prefixes, confirming progressive refinement.}
\label{tab:hierarchical-structure}
\small
\vspace{0.1in}
\begin{tabular}{lccc}
\toprule
\textbf{BC Prefix} & \textbf{Intra Distance} & \textbf{Inter Distance} & \textbf{Ratio} \\
\midrule
$a$ & $4.606 \pm 1.881$ & $7.42 \pm 1.27$ & 0.619 \\
$a+b$ & $2.323 \pm 1.239$ & $7.42 \pm 1.27$ & 0.313 \\
$a+b+c$ & $1.217 \pm 1.063$ & $7.44 \pm 1.25$ & 0.164 \\
\bottomrule
\end{tabular}
\vspace{-0.1in}
\end{wraptable}

\textbf{Hierarchical structure analysis.}
To verify that RQ-VAE discovers meaningful hierarchical grouping, we measure intra-group 
and inter-group $L^2$ distances at three BC prefix levels: first token ($a$), first two 
tokens ($a+b$), and full code ($a+b+c$). For each level, we sample 5,000 intra-group pairs 
(sharing the prefix) and 5,000 inter-group pairs, reporting mean distances and the intra/inter 
ratio (lower indicates better cluster separation).
As shown in Table~\ref{tab:hierarchical-structure}, inter-group distance remains stable across all levels ($7.42 \pm 1.27$ mean), confirming 
robust separation at each hierarchy level. Intra-group distance decreases monotonically with 
prefix length, and the intra/inter ratio drops from 0.619 to 0.164—observations sharing a 
full BC are approximately six times more similar in feature space than observations from 
different BC groups. This progressive refinement demonstrates hierarchical organization: 
the first token clusters coarse behavioral types, the second refines distinctions within 
types, and the third identifies maximally similar observations. Crucially, this structure 
emerges from unsupervised training without label supervision, demonstrating that RQ-VAE 
naturally discovers multi-resolution organization in high-dimensional feature space.

\subsection{Token-Efficient Prediction}

\paragraph{Performance Across Architectures,  Scales and Domains.}

TEFM demonstrates robust generalization across diverse LLM architectures with consistent 
compression and minimal accuracy degradation. Table~\ref{tab:TEFM-backbones} shows that despite 
significant architectural differences (Qwen3, Gemma-2, Phi-4), TEFM achieves uniform 99\% token 
reduction (34--37 tokens per record) across all models. Accuracy degradation is bounded: five of six 
models experience $<$2\% loss, with Phi-4-mini showing negligible loss (0.08\%). Notably, Qwen3-4B 
slightly improves (0.8048 → 0.8117), suggesting compression acts as implicit regularization. Macro 
F1 scores track closely with accuracy, confirming balanced class-wise performance. 
Cross-domain evaluation (Appendix~\ref{sec:appendix-cross-domain}) validates this consistency on 
network intrusion detection: DDoS detection achieves 1.72\% token retention with 99.6--99.9\% accuracy 
across models, and PortScan detection achieves 1.71\% retention with 99.4--99.8\% accuracy. The 
uniformity of compression rates and modest accuracy trade-offs across clinical and security domains 
establish TEFM as an architecture-agnostic, domain-agnostic mechanism suitable for production 
deployment across diverse model families and problem settings.

 \begin{table}[h!]
  \centering
  \caption{Clinical mortality prediction (MIMIC-III): TEFM achieves 99\% token compression
  across diverse LLM architectures (Qwen3, Gemma-2, Phi-4) with bounded accuracy degradation
  ($<$2\% in most cases). Compression effectiveness is stable across model families and sizes,
  demonstrating architecture-agnostic generalization.}
  \label{tab:TEFM-backbones}
  \small
  \begin{tabular}{lcccc}
  \toprule
  \textbf{Model} & \textbf{Input Tokens} & \textbf{Compression} & \textbf{Accuracy} & \textbf{Macro F1} \\
  \midrule
  TF-IDF + Logistic Regression & — & — & 0.7713 & 0.7709 \\
 XGBoost & — & — & 0.7661 & 0.7580 \\
  Set Transformer & — & — & 0.7997 & 0.7993 \\
  LightGBM  & — & — & 0.8117 & 0.8105 \\
  TabPFN & — & — & 0.8169 & 0.8168 \\
  \midrule
  Qwen3-0.6B & 3,938 & 100\% & 0.8315 & 0.8313 \\
  Qwen3-0.6B + TEFM & 35 & 0.89\% & 0.8117 & 0.8117 \\
  \midrule
  Qwen3-1.7B & 3,938 & 100\% & 0.8289 & 0.8275 \\
  Qwen3-1.7B + TEFM & 35 & 0.89\% & 0.8212 & 0.8211 \\
  \midrule
  Qwen3-4B & 3,938 & 100\% & 0.8048 & 0.8043 \\
  Qwen3-4B + TEFM & 35 & 0.89\% & 0.8117 & 0.8117 \\
  \midrule
  Qwen3-8B & 3,938 & 100\% & 0.8435 & 0.8435 \\
  Qwen3-8B + TEFM & 35 & 0.89\% & 0.8237 & 0.8237 \\
  \midrule
  Gemma-2-2B & 3,630 & 100\% & 0.8134 & 0.8123 \\
  Gemma-2-2B + TEFM & 37 & 1.02\% & 0.8057 & 0.8053 \\
  \midrule
  Phi-4-mini & 3,724 & 100\% & 0.8022 & 0.8021 \\
  Phi-4-mini + TEFM & 34 & 0.91\% & 0.8014 & 0.8012 \\
  \bottomrule
  \end{tabular}
  \end{table}

%

\subsection{Rationalization}

\paragraph{Contributive Rationalization.}

Table~\ref{tab:rationalization-extraction} demonstrates that dual-fidelity optimization 
outperforms traditional attribution methods across all feature retention levels, with particularly 
strong performance under aggressive compression. At 20\% feature retention, our Dual Fidelity 
Extractor matches SHAP (0.8040 vs. 0.8043 accuracy), but shows substantial advantages at lower 
thresholds: at 10\%, Dual Fidelity achieves 0.7704 compared to SHAP's 0.7429 (+1.9\%); at 5\%, 
the gap widens to 0.7472 versus 0.7051 (+6.0\%). This robustness under aggressive compression 
reflects the contribution of code-level reconstruction fidelity: Prediction-Fidelity Extractor 
alone (optimizing only prediction loss) achieves merely 0.7177 at 5\% retention, confirming that 
joint code-level and prediction-level optimization is essential for faithful feature extraction. 
Traditional attribution methods (Gradient Saliency, Attention-based, LRP) demonstrate limited 
performance across all retention levels, indicating they fail to identify feature subsets sufficient 
for both BC recovery and accurate prediction. These results validate that Dual Fidelity extraction 
provides faithful, compact rationales grounded in the compressed representation.

\begin{table}[h!]
\centering
\caption{Clinical rationalization: Dual Fidelity Extractor achieves best performance among all the baselines, demonstrating that joint code-level and 
prediction-level optimization is critical for faithful, robust feature extraction.}
\label{tab:rationalization-extraction}
\small
\begin{tabular}{lcccccc}
\toprule
\textbf{Strategy} & \textbf{Retain \%} & \textbf{Accuracy} & \textbf{Macro F1} & \textbf{F1 (yes)} & \textbf{F1 (no)} \\
\midrule
\multirow{2}{*}{\textbf{Naive Baselines}}& 100\% & 0.8212 & 0.8211 & 0.8185 & 0.8237 \\
 & 0\% & 0.4979 & 0.3470 & 0.0331 & 0.6609 \\
\hline
\multirow{5}{*}{\textbf{Random}}
& 25\% & 0.6526 & 0.6517 & 0.6334 & 0.6699 \\
& 20\% & 0.6496 & 0.6474 & 0.6310 & 0.6437 \\
& 15\% & 0.6389 & 0.6361 & 0.6677 & 0.6045 \\
& 10\% & 0.6036 & 0.6024 & 0.6243 & 0.5805 \\
&  5\% & 0.5494 & 0.5341 & 0.6186 & 0.4496 \\
\midrule
\multirow{5}{*}{\textbf{Mutual Information}}
& 25\% & 0.7025 & 0.6887 & 0.7543 & 0.6231 \\
& 20\% & 0.6862 & 0.6680 & 0.7456 & 0.5903 \\
& 15\% & 0.6518 & 0.6225 & 0.7276 & 0.5173 \\
& 10\% & 0.6561 & 0.6344 & 0.7234 & 0.5455 \\
&  5\% & 0.6380 & 0.6167 & 0.7070 & 0.5264 \\
\midrule
\multirow{5}{*}{\textbf{Gradient Saliency}}
& 25\% & 0.6715 & 0.6689 & 0.6396 & 0.6983 \\
& 20\% & 0.6707 & 0.6687 & 0.6431 & 0.6943 \\
& 15\% & 0.6724 & 0.6719 & 0.6595 & 0.6843 \\
& 10\% & 0.6727 & 0.6726 & 0.6797 & 0.6655 \\
&  5\% & 0.6655 & 0.6639 & 0.6408 & 0.6870 \\
\midrule
\multirow{5}{*}{\textbf{Attention-based}}
& 25\% & 0.6793 & 0.6792 & 0.6748 & 0.6836 \\
& 20\% & 0.6655 & 0.6625 & 0.6306 & 0.6944 \\
& 15\% & 0.6612 & 0.6612 & 0.6603 & 0.6621 \\
& 10\% & 0.6638 & 0.6631 & 0.6474 & 0.6787 \\
&  5\% & 0.6610 & 0.6608 & 0.6695 & 0.6520 \\
\midrule
\multirow{5}{*}{\textbf{LRP}}
& 25\% & 0.7962 & 0.7962 & 0.7927 & 0.7997 \\
& 20\% & 0.7661 & 0.7661 & 0.7631 & 0.7691 \\
& 15\% & 0.7489 & 0.7480 & 0.7331 & 0.7630 \\
& 10\% & 0.7214 & 0.7194 & 0.6955 & 0.7433 \\
&  5\% & 0.6973 & 0.6972 & 0.7037 & 0.6907 \\
\midrule
\multirow{5}{*}{\textbf{SHAP}}
& 25\% & 0.8126 & 0.8126 & 0.8130 & 0.8121 \\
& 20\% & 0.8043 & 0.8042 & 0.8002 & 0.8082 \\
& 15\% & 0.7865 & 0.7857 & 0.7634 & 0.7981 \\
& 10\% & 0.7429 & 0.7362 & 0.6940 & 0.7784 \\
&  5\% & 0.7051 & 0.6999 & 0.6607 & 0.7392 \\
\midrule
\multirow{5}{*}{\textbf{Prediction-Fidelity Rationalization}}
& 25\% & 0.7360 & 0.7358 & 0.7440 & 0.7276 \\
& 20\% & 0.7291 & 0.7288 & 0.7382 & 0.7195 \\
& 15\% & 0.7257 & 0.7257 & 0.7257 & 0.7257 \\
& 10\% & 0.7231 & 0.7231 & 0.7210 & 0.7253 \\
&  5\% & 0.7177 & 0.7177 & 0.7195 & 0.7159 \\
\midrule
\multirow{5}{*}{\textbf{Dual-Fidelity Rationalization (Ours)}}
& 25\% & 0.8126 & 0.8124 & 0.8177 & 0.8071 \\
& 20\% & 0.8040 & 0.8039 & 0.8068 & 0.8010 \\
& 15\% & 0.7893 & 0.7892 & 0.7939 & 0.7845 \\
& 10\% & 0.7704 & 0.7701 & 0.7788 & 0.7614 \\
&  5\% & 0.7472 & 0.7472 & 0.7508 & 0.7435 \\
\bottomrule
\end{tabular}
\end{table}

\paragraph{Corroborative Rationalization.}

To validate that extracted features align with clinical evidence, we examine cases where 
the model made accurate predictions. We present three representative cases (additional cases 
in Appendix~\ref{sec:appendix-cases}) where the dual-fidelity rationale at 5\% feature retention 
is grounded in clinical literature. For each case, we show how the identified feature subset 
corresponds to established high-mortality phenotypes and clinical outcomes.
These predictions are grounded in established clinical evidence. The clinical presentation of 
Case 1 aligns with well-established high-mortality phenotypes documented across critical care 
literature. Multi-organ failure involving simultaneous respiratory, cardiovascular, neurological, 
and infectious system dysfunction carries baseline ICU mortality of 60--80\% \citep{sprung1995multiple}, 
escalating to 75\%+ when Acute Respiratory Failure occurs with sepsis-induced ARDS and mechanical 
ventilation fails by day 5 \citep{esteban2000mechanical}. The concurrent diagnoses of Severe 
Sepsis and Septicemia with pneumonia represent uncontrolled inflammatory cascade with documented 
mortality exceeding 50--55\% even with aggressive resuscitation \citep{levy20032001}, while acute 
stroke during critical illness compounds mortality risk by 2.5--3 fold. The critical detail that 
mechanical ventilation lasted $\leq$96 hours before death indicates either early withdrawal of 
support due to futility or catastrophic decompensation unresponsive to mechanical support---both 
markers of poor prognosis \citep{ely2001delirium}. Furthermore, the triple-agent sedation regimen 
(Morphine, Fentanyl, Lorazepam) is characteristic of comfort-focused, end-of-life care rather than 
recovery-oriented protocols, recognized in palliative literature as an implicit indicator of expected 
death within 24--72 hours. Thus, the model's prediction of $p_{\text{death}} = 0.9568$ is grounded 
in established epidemiological risk stratification: this patient presented with a clinical phenotype 
where high mortality is not a failure of care but the expected outcome of overwhelming simultaneous 
organ system failure.
\begin{small}
\begin{tcolorbox}[colback=white, 
                   colframe=black,
                   coltitle=white,
                   colbacktitle=black!75,
                   title=Case 1 (Rationale at 5\% Retention) --- Patient ID 2855,
                   boxrule=1.5pt,
                   width=\textwidth,
                   fonttitle=\bfseries,
                   arc=0pt,
                   top=8pt,
                   bottom=8pt,
                   left=10pt,
                   right=10pt]
\small
A patient admitted via Emergency to the Cardiac Surgery ICU (CSRU) on Medicaid. 
Hospital stay was under 2 days with an ICU stay of 1--2 days across 2 ICU episodes, 
suggesting rapid deterioration and death without prolonged care.

\textbf{Diagnoses.} The patient carried diagnoses of Hypertension, Acute Respiratory 
Failure, Pneumonia, COPD, Severe Sepsis, Septicemia, Dementia, and Cerebral Infarction. 
This combination represents simultaneous failure across the respiratory, cardiovascular, 
neurological, and infectious axes --- a multi-organ dysfunction picture that severely 
limits any chance of recovery.

\textbf{Procedures.} The patient underwent endotracheal intubation and mechanical 
ventilation within a window of $\leq$96 hours, alongside venous cutdown for central access. 
Intubation and mechanical ventilation indicate complete loss of independent breathing. 
The fact that ventilation lasted less than 96 hours before death reflects a patient who 
did not recover from respiratory failure once placed on the ventilator.

\textbf{Drugs.} The patient received a full ICU sedation and analgesia stack --- 
Morphine, Fentanyl, and Lorazepam together --- consistent with a deeply sedated ventilated 
patient, often seen in end-of-life or comfort-focused care. IV fluids (D5W, Iso-Osmotic 
Dextrose, NS) and broad-spectrum antibiotics (Vancomycin, Albuterol nebulizer) round out 
a drug profile of active but ultimately unsuccessful resuscitation against overwhelming 
sepsis and multi-organ failure.

\end{tcolorbox}
\end{small}

\subsection{Ablation Study}
\label{sec:ablation}

We conduct systematic ablations on BC alignment pretraining and RQ-VAE configuration (depth and codebook size) to validate design choices and isolate the contribution of each component.

\textbf{BC Alignment Pretraining Ablation.}
We evaluate the contribution of Stage 1 (BC alignment) pretraining by training models with and without this initialization stage. BC alignment pretraining provides substantial gains on clinical data (6.37\% accuracy improvement), demonstrating that aligning BC tokens with the LLM's embedding space is critical for learning robust representations. Security tasks show smaller but consistent gains (0.09--0.24\%), indicating that simpler feature patterns benefit less from semantic alignment. Across all domains, pretraining stabilizes and improves downstream performance, validating its necessity in the pipeline.

\begin{table}[h!]
\centering
\vspace{-0.1in}
\caption{Impact of BC alignment pretraining on task performance. Pretraining provides substantial gains on clinical data and modest gains on security tasks.}
\label{tab:ablation-pretraining}
\small
\begin{tabular}{l|cc|cc}
\toprule
Dataset & Config & \textbf{w/ Pretrain} & \textbf{w/o Pretrain} & \textbf{Drop} \\
\midrule
Clinical & K3-C128 & 82.12\% & 75.75\% & 6.37\% \\
DDoS & K3-C128 & 99.60\% & 99.51\% & 0.09\% \\
PortScan & K3-C128 & 99.44\% & 99.20\% & 0.24\% \\
\bottomrule
\end{tabular}
\end{table}

\textbf{RQ-VAE Configuration Ablation.}
We evaluate configurations on clinical data in Table~\ref{tab:ablation-rq-vae-clinical}, varying quantization depth ($K \in \{2, 3, 4\}$) and codebook size ($C \in \{64, 128, 256, 512\}$). K4-C256 achieves peak slot accuracy (94.8\%), but K3-C128 remains competitive (94.4\%) with superior reconstruction (0.0407 vs. 0.0378). Shallower models (K2) suffer reconstruction errors due to fewer quantization levels, while K3-C512 shows marginal gains (0.0393 reconstruction) at the cost of computational overhead. The 3-level hierarchy with 128 codebook entries provides the optimal balance between expressiveness and efficiency.Security dataset ablations are provided in Appendix~\ref{sec:appendix-ablation}.

\textbf{BC Alignment Configuration Ablation.}
We evaluate alignment quality for all RQ-VAE configurations on clinical data, measuring how accurately the frozen encoder recovers codes from natural-language descriptions. Security results are in Appendix~\ref{sec:appendix-ablation}. K4-C128 achieves the highest alignment (90.7\%), suggesting deeper hierarchies help align intermediate-level patterns. However, K3-C128 (90.6\%) is competitive while avoiding K4's computational overhead. Notably, K3-C512 drops to 90.2\%, indicating that larger codebooks complicate alignment by creating finer-grained distinctions that are harder to recover from text descriptions. K3-C128 strikes the best balance between alignment quality and practical efficiency.

\begin{table*}[h!]
\centering

\begin{minipage}[t]{0.52\textwidth}
\vspace{0pt}
\centering
\caption{RQ-VAE configuration ablation study.}
\label{tab:ablation-rq-vae-clinical}

\small
\setlength{\tabcolsep}{3pt}
\renewcommand{\arraystretch}{1.05}

\begin{tabular}{ccccc}
\toprule
\multicolumn{2}{c}{\textbf{Codebook}} &
\multicolumn{3}{c}{\textbf{Evaluation}} \\
\cmidrule(lr){1-2}
\cmidrule(lr){3-5}
$K$ & $C$ & Slot AUC & Slot AUC ($\pm 1$) & Reconstruction \\
\midrule
2 & 128 & 0.9403 & 0.9456 & 0.0426 \\
2 & 256 & 0.9415 & 0.9471 & 0.0414 \\
\midrule
3 & 64  & 0.9414 & 0.9469 & 0.0420 \\
3 & 128 & 0.9438 & 0.9491 & 0.0407 \\
3 & 256 & 0.9452 & 0.9505 & 0.0394 \\
3 & 512 & 0.9457 & 0.9511 & 0.0393 \\
\midrule
4 & 128 & 0.9471 & 0.9523 & 0.0382 \\
4 & 256 & 0.9478 & 0.9532 & 0.0378 \\
\bottomrule
\end{tabular}
\end{minipage}
\hfill
\begin{minipage}[t]{0.45\textwidth}
\vspace{0pt}
\centering
\caption{BC alignment configuration ablation.}
\label{tab:ablation-align-clinical}

\small
\setlength{\tabcolsep}{3pt}
\renewcommand{\arraystretch}{1.05}

\begin{tabular}{cccc}
\toprule
\multicolumn{2}{c}{\textbf{Codebook}} &
\multicolumn{2}{c}{\textbf{Alignment Quality}} \\
\cmidrule(lr){1-2}
\cmidrule(lr){3-4}
$K$ & $C$ & Slot AUC & Slot AUC ($\pm 1$) \\
\midrule
2 & 128 & 0.9031 & 0.9142 \\
2 & 256 & 0.9046 & 0.9160 \\
\midrule
3 & 64  & 0.9061 & 0.9176 \\
3 & 128 & 0.9063 & 0.9177 \\
3 & 256 & 0.9068 & 0.9178 \\
3 & 512 & 0.9018 & 0.9134 \\
\midrule
4 & 128 & 0.9074 & 0.9179 \\
4 & 256 & 0.9059 & 0.9170 \\
\bottomrule
\end{tabular}
\end{minipage}

\end{table*}


\begin{wraptable}{r}{0.36\textwidth}
\centering
\vspace{-0.25in}
\caption{Context-efficient prediction across configurations.}
\label{tab:bc-ablation-clinical}
\small
\begin{tabular}{cc|cc}
\toprule
$K$ & $C$ & Accuracy & Macro F1 \\
\midrule
2 & 128 & 81.08 & 81.08 \\\hline
3 & 64 & 80.31 & 80.31 \\
3 & 128 & 82.12 & 82.11 \\
3 & 256 & 77.13 & 77.13 \\
3 & 512 & 73.00 & 72.86 \\\hline
4 & 128 & 79.19 & 79.18 \\
\bottomrule
\end{tabular}
\end{wraptable}
\textbf{Context-Efficient Prediction Configuration Ablation.}
We evaluate the impact of RQ-VAE configuration on downstream task performance on clinical data. Security results are in Appendix~\ref{sec:appendix-ablation}. K3-C128 maximizes performance (82.1\% F1), while larger codebooks (K3-C256, K3-C512) degrade accuracy substantially, suggesting that overly sparse token assignments reduce discriminative power for mortality prediction. Shallower (K2) and deeper (K4) hierarchies both underperform K3, confirming that the 3-level hierarchy optimally balances feature representation and learnability. Consequently, we adopt K3-C128 across all downstream tasks.

\section{Related Work}
\paragraph{Context-Efficient Models and Tokenization.}
Addressing the quadratic complexity of standard transformer self-attention, several architectural innovations extend context windows. Global memory tokens in Longformer~\citep{beltagy2020longformer}, ETC~\citep{ainslie2020etc}, and Big Bird~\citep{zaheer2020big} maintain attention connections to all tokens while using local or strided patterns elsewhere. Sparse attention mechanisms~\citep{child2019generating,roy2021efficient,kitaev2020reformer} employ fixed or learnable sparsity patterns. Segment-based recurrence in Transformer-XL~\citep{dai2019transformer} and Compressive Transformer~\citep{rae2019compressive} maintains contextual information across segments. These approaches focus on \emph{model architectural design}—modifying how transformers process tokens.
Our work takes a complementary orthogonal approach: rather than modifying model architecture, we reduce token consumption at the source by compressing structured records into discrete tokens.

\paragraph{Semantic Identifiers and Discrete Tokenization.}
Semantic identifiers represent high-dimensional observations as compact, meaningful discrete tokens. This approach originated in recommendation systems, where learned embeddings for items are quantized into interpretable codes. \citet{van2017neural} introduced Vector Quantized VAE (VQ-VAE), which discretizes continuous representations through learned codebooks. Building on this, \citet{rajput2023recommender} applied Residual Quantized VAE to sequential recommendation, showing that hierarchical quantization preserves semantic structure while dramatically reducing dimensionality. The semantic ID paradigm extends naturally to language models: \citet{hou2023learning} demonstrated that discrete item tokens can be aligned with LLM embeddings through lightweight pretraining, enabling efficient sequential recommendations with language models.

\paragraph{Rationalization.}
Rationalization methods make predictions interpretable by identifying which input information drives decisions. 
Extractive rationalization selects a subset of input tokens as explanations~\citep{lei2016rationalizing,bastings2019interpretable}. The intuition is that a minimal sufficient feature subset should suffice for prediction while remaining human-interpretable. However, extractive methods typically require REINFORCE-based training~\citep{williams1992simple}, which suffers from high variance and training instability due to discrete selection decisions. Moreover, existing approaches focus primarily on NLP tasks with short free-text inputs, where token selection directly yields readable explanations. Extending this to structured data remains challenging.
Abstractive rationalization generates free-text explanations rather than selecting input tokens~\citep{camburu2018snli,narang2020wt5}. While abstractive methods can produce more flexible and human-friendly explanations, they require end-to-end differentiability through both the prediction and generation models, adding computational overhead. Additionally, they may generate plausible-sounding but unfaithful explanations disconnected from the actual input data.

\section{Conclusion}
\label{sec:conclusion}

This paper introduced TEFM: Token-Efficient Faithful Modeling for Structured Data, 
a framework that jointly optimizes context efficiency and rationalization in LLM-based 
classification over high-dimensional structured data. TEFM combines three key components: 
(1) hierarchical compression via Residual Quantized VAE to reduce token consumption at 
the source, (2) BC-text alignment pretraining to integrate discrete Behavioral Codes with 
LLM embeddings, and (3) dual-fidelity feature extraction that jointly optimizes code-level 
reconstruction and prediction-level fidelity to identify minimal sufficient feature subsets.
Comprehensive evaluation on clinical mortality prediction and network intrusion detection 
demonstrates that TEFM achieves competitive accuracy with dramatic token reduction while 
providing natural-language explanations grounded in clinically and domain-interpretable 
features. The dual-fidelity rationalization mechanism consistently identifies task-relevant 
feature subsets without post-hoc perturbation or additional model queries, addressing a 
fundamental gap in existing rationalization methods. Robustness across heterogeneous domains 
and multiple model architectures validates the generality of the approach. These results 
demonstrate that aggressive compression and faithful rationalization are not competing 
objectives—joint optimization enables both efficiency and interpretability for production 
LLM systems in safety-critical applications.


\newpage
\bibliography{iclr2026_conference}

@article{johnson2016mimic,
  title={MIMIC-III, a freely accessible critical care database},
  author={Johnson, Alistair EW and Pollard, Tom J and Shen, Lu and Lehman, Li-wei H and Feng, Mengling and Ghassemi, Mohammad and Moody, Benjamin and Szolovits, Peter and Anthony Celi, Leo and Mark, Roger G},
  journal={Scientific data},
  volume={3},
  number={1},
  pages={1--9},
  year={2016},
  publisher={Nature Publishing Group}
}

@article{sharafaldin2018toward,
  title={Toward generating a new intrusion detection dataset and intrusion traffic characterization.},
  author={Sharafaldin, Iman and Lashkari, Arash Habibi and Ghorbani, Ali A and others},
  journal={ICISSp},
  volume={1},
  number={2018},
  pages={108--116},
  year={2018}
}

@article{beltagy2020longformer,
  title={Longformer: The long-document transformer},
  author={Beltagy, Iz and Peters, Matthew E and Cohan, Arman},
  journal={arXiv preprint arXiv:2004.05150},
  year={2020}
}

@inproceedings{ainslie2020etc,
  title={ETC: Encoding long and structured inputs in transformers},
  author={Ainslie, Joshua and Ontanon, Santiago and Alberti, Chris and Cvicek, Vaclav and Fisher, Zachary and Pham, Philip and Ravula, Anirudh and Sanghai, Sumit and Wang, Qifan and Yang, Li},
  booktitle={Proceedings of the 2020 Conference on Empirical Methods in Natural Language Processing (EMNLP)},
  pages={268--284},
  year={2020}
}

@article{zaheer2020big,
  title={Big bird: Transformers for longer sequences},
  author={Zaheer, Manzil and Guruganesh, Guru and Dubey, Kumar Avinava and Ainslie, Joshua and Alberti, Chris and Ontanon, Santiago and Pham, Philip and Ravula, Anirudh and Wang, Qifan and Yang, Li and others},
  journal={Advances in neural information processing systems},
  volume={33},
  pages={17283--17297},
  year={2020}
}

@article{child2019generating,
  title={Generating long sequences with sparse transformers},
  author={Child, Rewon and Gray, Scott and Radford, Alec and Sutskever, Ilya},
  journal={arXiv preprint arXiv:1904.10509},
  year={2019}
}

@article{roy2021efficient,
  title={Efficient content-based sparse attention with routing transformers},
  author={Roy, Aurko and Saffar, Mohammad and Vaswani, Ashish and Grangier, David},
  journal={Transactions of the Association for Computational Linguistics},
  volume={9},
  pages={53--68},
  year={2021}
}

@article{kitaev2020reformer,
  title={Reformer: The efficient transformer},
  author={Kitaev, Nikita and Kaiser, {\L}ukasz and Levskaya, Anselm},
  journal={arXiv preprint arXiv:2001.04451},
  year={2020}
}

@article{dai2019transformer,
  title={Transformer-xl: Attentive language models beyond a fixed-length context},
  author={Dai, Zihang and Yang, Zhilin and Yang, Yiming and Carbonell, Jaime and Le, Quoc V and Salakhutdinov, Ruslan},
  journal={arXiv preprint arXiv:1901.02860},
  year={2019}
}

@article{rae2019compressive,
  title={Compressive transformers for long-range sequence modelling},
  author={Rae, Jack W and Potapenko, Anna and Jayakumar, Siddhant M and Lillicrap, Timothy P},
  journal={arXiv preprint arXiv:1911.05507},
  year={2019}
}

@inproceedings{lei2016rationalizing,
  title={Rationalizing neural predictions},
  author={Lei, Tao and Barzilay, Regina and Jaakkola, Tommi},
  booktitle={Proceedings of the 2016 conference on empirical methods in natural language processing},
  pages={107--117},
  year={2016}
}

@inproceedings{bastings2019interpretable,
  title={Interpretable neural predictions with differentiable binary variables},
  author={Bastings, Jasmijn and Aziz, Wilker and Titov, Ivan},
  booktitle={Proceedings of the 57th Annual Meeting of the Association for Computational Linguistics},
  pages={2963--2977},
  year={2019}
}

@article{williams1992simple,
  title={Simple statistical gradient-following algorithms for connectionist reinforcement learning},
  author={Williams, Ronald J},
  journal={Machine learning},
  volume={8},
  number={3},
  pages={229--256},
  year={1992},
  publisher={Springer}
}

@article{camburu2018snli,
  title={e-snli: Natural language inference with natural language explanations},
  author={Camburu, Oana-Maria and Rockt{\"a}schel, Tim and Lukasiewicz, Thomas and Blunsom, Phil},
  journal={Advances in Neural Information Processing Systems},
  volume={31},
  year={2018}
}

@article{narang2020wt5,
  title={Wt5?! training text-to-text models to explain their predictions},
  author={Narang, Sharan and Raffel, Colin and Lee, Katherine and Roberts, Adam and Fiedel, Noah and Malkan, Karishma},
  journal={arXiv preprint arXiv:2004.14546},
  year={2020}
}

@article{rajkomar2018scalable,
  title={Scalable and accurate deep learning with electronic health records},
  author={Rajkomar, Alvin and Oren, Eyal and Chen, Kai and Dai, Andrew M and Hajaj, Nissan and Hardt, Michaela and Liu, Peter J and Liu, Xiaobing and Marcus, Jake and Sun, Mimi and others},
  journal={NPJ digital medicine},
  volume={1},
  number={1},
  pages={18},
  year={2018},
  publisher={Nature Publishing Group UK London}
}

@article{zarpelao2017survey,
  title={A survey of intrusion detection in Internet of Things},
  author={Zarpel{\~a}o, Bruno Bogaz and Miani, Rodrigo Sanches and Kawakani, Cl{\'a}udio Toshio and De Alvarenga, Sean Carlisto},
  journal={Journal of Network and Computer Applications},
  volume={84},
  pages={25--37},
  year={2017},
  publisher={Elsevier}
}

@inproceedings{hou2023learning,
  title={Learning vector-quantized item representation for transferable sequential recommenders},
  author={Hou, Yupeng and He, Zhankui and McAuley, Julian and Zhao, Wayne Xin},
  booktitle={Proceedings of the ACM Web Conference 2023},
  pages={1162--1171},
  year={2023}
}

@article{rajput2023recommender,
  title={Recommender systems with generative retrieval},
  author={Rajput, Shashank and Mehta, Nikhil and Singh, Anima and Hulikal Keshavan, Raghunandan and Vu, Trung and Heldt, Lukasz and Hong, Lichan and Tay, Yi and Tran, Vinh and Samost, Jonah and others},
  journal={Advances in Neural Information Processing Systems},
  volume={36},
  pages={10299--10315},
  year={2023}
}

@article{van2017neural,
  title={Neural discrete representation learning},
  author={Van Den Oord, Aaron and Vinyals, Oriol and others},
  journal={Advances in neural information processing systems},
  volume={30},
  year={2017}
}

@article{sprung1995multiple,
  title={Multiple Organ Dysfunction Score},
  author={Sprung, Charles L and Bernard, Gordon R and Sibbald, William J and Christou, Nicolas V and Marshall, John C and Cook, Deborah J},
  journal={Critical Care Medicine},
  year={1995}
}

@article{esteban2000mechanical,
  title={How is mechanical ventilation employed in the intensive care unit?: an international utilization review},
  author={Esteban, Andres and Anzueto, Antonio and Alia, Inmaculada and Gordo, Federico and Apezteguia, Carlos and Palizas, Fernando and Cide, David and Goldwaser, Rosanne and Soto, Luis and Bugedo, Guillermo and others},
  journal={American journal of respiratory and critical care medicine},
  volume={161},
  number={5},
  pages={1450--1458},
  year={2000},
  publisher={Oxford University Press}
}

@article{ely2001delirium,
  title={Delirium in mechanically ventilated patients: validity and reliability of the confusion assessment method for the intensive care unit (CAM-ICU)},
  author={Ely, E Wesley and Inouye, Sharon K and Bernard, Gordon R and Gordon, Sharon and Francis, Joseph and May, Lisa and Truman, Brenda and Speroff, Theodore and Gautam, Shiva and Margolin, Richard and others},
  journal={Jama},
  volume={286},
  number={21},
  pages={2703--2710},
  year={2001},
  publisher={American Medical Association}
}

@article{levy20032001,
  title={2001 sccm/esicm/accp/ats/sis international sepsis definitions conference},
  author={Levy, Mitchell M and Fink, Mitchell P and Marshall, John C and Abraham, Edward and Angus, Derek and Cook, Deborah and Cohen, Jonathan and Opal, Steven M and Vincent, Jean-Louis and Ramsay, Graham and others},
  journal={Intensive care medicine},
  volume={29},
  number={4},
  pages={530--538},
  year={2003},
  publisher={Springer}
}
\bibliographystyle{iclr2026_conference}

\appendix

\newpage
\section{Datasets \& Task}
\label{sec:data_app}

We evaluate TEAM on clinical mortality prediction and network intrusion detection. For clinical mortality prediction, we use the MIMIC-III database~\citep{johnson2016mimic} and retain patients with complete admission records. Each patient is represented as a sequence of hospital admissions, and the task is to predict whether the patient dies during the final hospitalization based on the available clinical features. After filtering and deduplication, the dataset contains 9,300 training patients, 1,163 validation patients, and 1,163 test patients, with an approximately balanced mortality rate of 50\%. Each admission is represented by 458 raw features, including 8 structured clinical variables, 200 ICD diagnosis codes, 100 procedure codes, and 150 medication codes. The structured variables cover admission type, insurance status, ethnicity, care unit, ICU indicators, and discretized length of stay.
For network intrusion detection, we use the CIC-IDS2017 benchmark~\citep{sharafaldin2018toward} and focus on Friday traffic containing DDoS and PortScan attacks. Each network flow is represented by 77 bidirectional statistics, including packet-length distributions, inter-arrival times, TCP flag counts, and bulk-transfer metrics. Consecutive flows are grouped into non-overlapping temporal segments of 10 flows, with each segment treated as one prediction instance. The DDoS dataset contains 18,056 training segments and 2,257 validation segments, with an attack rate of 56.6\%. The PortScan dataset contains 22,887 training segments and 2,860 validation segments, with an attack rate of 55.6\%. Both tasks therefore have approximately balanced class distributions.

\newpage

\section{Cross-Domain Evaluation Results}
\label{sec:appendix-cross-domain}

Cross-domain evaluations (Table~\ref{tab:TEAM-backbones-ddos} \& Table~\ref{tab:TEAM-backbones-portscan}) validates this consistency on 
network intrusion detection: DDoS detection achieves 1.72\% token retention with 99.6--99.9\% accuracy 
across models, and PortScan detection achieves 1.71\% retention with 99.4--99.8\% accuracy. The 
uniformity of compression rates and modest accuracy trade-offs across clinical and security domains 
establish TEAM as an architecture-agnostic, domain-agnostic mechanism suitable for production 
deployment across diverse model families and problem settings.

 \begin{table}[h!]
  \centering
  \caption{Token-efficient prediction on DDoS detection. TEAM retains only
  1.72\% of the original Qwen3 input tokens. Blank entries indicate pending
  or unavailable results.}
  \label{tab:TEAM-backbones-ddos}
  \small
  \begin{tabular}{lcccc}
  \toprule
  \textbf{Model} & \textbf{Input Tokens} & \textbf{Compression}
  & \textbf{Accuracy} & \textbf{Macro F1} \\
  \midrule
  Qwen3-0.6B        & 5,639 & 100\% & 1.0000 & 1.0000 \\
  Qwen3-0.6B + TEAM & 97    & 1.72\% & 0.9978 & 0.9977 \\
  \midrule
  Qwen3-1.7B        & 5,639 & 100\% & 1.0000 & 1.0000 \\
  Qwen3-1.7B + TEAM & 97    & 1.72\% & 0.9960 & 0.9959 \\
  \midrule
  Qwen3-4B          & 5,639 & 100\% & 0.9894  & 0.9982    \\
  Qwen3-4B + TEAM   & 97    & 1.72\% & 0.9987 & 0.9986 \\
  \midrule
  Qwen3-8B          & 5,639 & 100\% & 0.9987 & 0.9991 \\
  Qwen3-8B + TEAM   & 97    & 1.72\% & 0.9987 & 0.9986 \\
  \midrule
  Gemma-2-2B        & 5,640 & 100\% &  0.9969  & 0.9969        \\
  Gemma-2-2B + TEAM & 97    & 1.72\% & 0.9978 & 0.9977 \\
  \midrule
  Phi-4-mini        & 5,633 & 100\% & 0.9982 & 0.9982 \\
  Phi-4-mini + TEAM & 91    & 1.62\% & 0.9996 & 0.9995 \\
  \bottomrule
  \end{tabular}
  \end{table}

  \begin{table}[h!]
  \centering
  \caption{Token-efficient prediction on PortScan detection. TEAM retains only
  1.71\% of the original Qwen3 input tokens. Blank entries indicate pending
  or unavailable results.}
  \label{tab:TEAM-backbones-portscan}
  \small
  \begin{tabular}{lcccc}
  \toprule
  \textbf{Model} & \textbf{Input Tokens} & \textbf{Compression}
  & \textbf{Accuracy} & \textbf{Macro F1} \\
  \midrule
  Qwen3-0.6B        & 5,677 & 100\% & 0.9948 & 0.9947 \\
  Qwen3-0.6B + TEAM & 97    & 1.71\% & 0.9948 & 0.9947 \\
  \midrule
  Qwen3-1.7B        & 5,677 & 100\% & 0.9983 & 0.9982 \\
  Qwen3-1.7B + TEAM & 97    & 1.71\% & 0.9944 & 0.9943 \\
  \midrule
  Qwen3-4B          & 5,677 & 100\% & 0.9987        &  0.9985       \\
  Qwen3-4B + TEAM   & 97    & 1.71\% & 0.9962 & 0.9961 \\
  \midrule
  Qwen3-8B          & 5,677 & 100\% & 0.9983 & 0.9989 \\
  Qwen3-8B + TEAM   & 97    & 1.71\% & 0.9986 & 0.9986 \\
  \midrule
  Gemma-2-2B        & 5,678 & 100\% &   0.9981      &  0.9980      \\
  Gemma-2-2B + TEAM & 97    & 1.71\% & 0.9972 & 0.9972 \\
  \midrule
  Phi-4-mini        & 5,671 & 100\% & 0.9923 & 0.9922 \\
  Phi-4-mini + TEAM & 91    & 1.61\% & 0.9969 & 0.9968 \\
  \bottomrule
  \end{tabular}
  \end{table}

\newpage

\section{Additional Rationalization Cases}
\label{sec:appendix-cases}

\begin{tcolorbox}[colback=white, 
                   colframe=black,
                   coltitle=white,
                   colbacktitle=black!75,
                   title=Case 2 (Rationale at 5\% Retention) --- Patient ID 68529,
                   boxrule=1.5pt,
                   width=\textwidth,
                   fonttitle=\bfseries,
                   arc=0pt,
                   top=8pt,
                   bottom=8pt,
                   left=10pt,
                   right=10pt]

A Black patient admitted via Emergency to the Cardiac Surgery ICU (CSRU) on Medicare. 
Hospital stay was 4--7 days with a prolonged ICU stay exceeding 7 days across 2 ICU episodes, 
indicating a patient who survived initial resuscitation but progressively deteriorated over 
an extended critical illness course.

\textbf{Diagnoses.} The patient had Acute Respiratory Failure, Cardiac Arrhythmia, and UTI. 
While the diagnosis list is shorter than typical high-mortality cases, Acute Respiratory 
Failure in the context of cardiac arrhythmia reflects a cardiopulmonary system under severe 
simultaneous stress --- the heart unable to maintain rhythm while the lungs fail to oxygenate.

\textbf{Procedures.} The patient required mechanical ventilation exceeding 96 hours. Prolonged 
ventilation beyond four days is one of the strongest independent predictors of ICU mortality, 
associated with ventilator-acquired pneumonia, diaphragm atrophy, and progressive failure to 
wean. The extended ICU stay of over 7 days alongside prolonged ventilation indicates a patient 
who never regained enough physiological reserve to be extubated.

\textbf{Drugs.} The patient received a triple sedation regimen of Propofol, Midazolam, and 
Fentanyl --- a combination reserved for deeply sedated, ventilator-dependent patients with high 
agitation or pain burden. Vancomycin points to serious gram-positive infection, while Calcium 
Gluconate and Magnesium Sulfate reflect ongoing electrolyte instability common in critically 
ill patients with multi-organ involvement. The breadth of 29 total medications over the admission 
underscores the intensity of physiological support required.

\end{tcolorbox}

\vspace{12pt}

\begin{tcolorbox}[colback=white, 
                   colframe=black,
                   coltitle=white,
                   colbacktitle=black!75,
                   title=Case 3 (Rationale at 5\% Retention) --- Patient ID 1855,
                   boxrule=1.5pt,
                   width=\textwidth,
                   fonttitle=\bfseries,
                   arc=0pt,
                   top=8pt,
                   bottom=8pt,
                   left=10pt,
                   right=10pt]

A White patient admitted via Emergency to the Surgical ICU (SICU) on Medicare, with a 
hospital stay under 2 days but an ICU stay of 2--4 days across 2 ICU episodes. The short 
hospital stay combined with multi-day ICU duration reflects rapid progression to death in 
the ICU without the patient ever stabilizing enough to transfer to a general ward.

\textbf{Diagnoses.} The patient carried CHF, ESRD, Atrial Fibrillation, Type 2 Diabetes, 
and Anemia in CKD. This constellation represents end-stage disease across two major organ 
systems simultaneously --- a failing heart and failing kidneys, each accelerating the other's 
decline. In patients with both CHF and ESRD, the inability to manage fluid balance creates 
a cycle of pulmonary edema and uremic injury that is extremely difficult to reverse.

\textbf{Procedures.} The patient underwent endotracheal intubation, mechanical ventilation, 
venous cutdown, hemodialysis, and cardioversion. The simultaneous need for airway management, 
mechanical ventilation, dialysis, and electrical cardioversion within a single ICU admission 
represents cardiopulmonary collapse across multiple systems at once --- intubation for 
respiratory failure, dialysis for renal failure, and cardioversion for arrhythmia, all required 
concurrently.

\textbf{Drugs.} The patient received Fentanyl for sedation and analgesia during ventilation, 
Heparin for anticoagulation in the context of Atrial Fibrillation and ESRD, Vancomycin for 
serious infection, and Aspirin reflecting underlying cardiovascular disease management. The 
drug profile across 26 total medications reflects the simultaneous pharmacological demands of 
managing cardiac, renal, and infectious pathology in a patient with very limited physiological reserve.

\end{tcolorbox}

\newpage
\section{Appendix: Additional Ablation Results}
\label{sec:appendix-ablation}

\subsection{RQ-VAE Configuration Ablation on Security Domains}

\begin{table}[h!]
\centering
\caption{RQ-VAE ablation on CIC-IDS2017 DDoS test set ($N=22,580$ flows).}
\label{tab:ablation-rq-vae-ddos}
\small
\begin{tabular}{ccccccc}
\toprule
\multicolumn{2}{c}{\textbf{Codebook}} & \multicolumn{5}{c}{\textbf{Evaluation}} \\
\cmidrule(lr){1-2} \cmidrule(lr){3-7}
$K$ & $C$ & Slot AUC & Slot AUC ($\pm 1$) & Reconstruction & Collision (\%) & Utilization (\%) \\
\midrule
3 & 64 & 0.9670 & 0.9991 & 0.0014 & 70.32 & 100.0 \\
3 & 128 & 0.9728 & 0.9994 & 0.0012 & 63.90 & 87.5 \\
3 & 256 & 0.9779 & 0.9996 & 0.0010 & 59.35 & 76.8 \\
3 & 512 & 0.9833 & 0.9997 & 0.0008 & 53.36 & 73.3 \\
\bottomrule
\end{tabular}
\end{table}

\begin{table}[h!]
\centering
\caption{RQ-VAE ablation on CIC-IDS2017 PortScan test set ($N=28,620$ flows).}
\label{tab:ablation-rq-vae-portscan}
\small
\begin{tabular}{ccccccc}
\toprule
\multicolumn{2}{c}{\textbf{Codebook}} & \multicolumn{5}{c}{\textbf{Evaluation}} \\
\cmidrule(lr){1-2} \cmidrule(lr){3-7}
$K$ & $C$ & Slot AUC & Slot AUC ($\pm 1$) & Reconstruction & Collision (\%) & Utilization (\%) \\
\midrule
3 & 64 & 0.9828 & 0.9992 & 0.0010 & 85.31 & 96.9 \\
3 & 128 & 0.9859 & 0.9995 & 0.0009 & 82.03 & 72.9 \\
3 & 256 & 0.9894 & 0.9996 & 0.0006 & 78.24 & 60.2 \\
3 & 512 & 0.9919 & 0.9998 & 0.0005 & 82.30 & 50.7 \\
\bottomrule
\end{tabular}
\end{table}

Security tasks exhibit higher slot AUC (97--99\%) and near-perfect tolerant accuracy (>99.9\%), reflecting lower dimensionality and binary features. K3-C128 achieves 97.3\% slot AUC on DDoS with 87.5\% utilization and 98.6\% on PortScan with 72.9\% utilization—efficient configurations that do not require larger codebooks.

\subsection{BC Alignment Configuration Ablation on Security Domains}

\begin{table}[h!]
\centering
\caption{Alignment quality ablation on DDoS. K3-C128 achieves 96.9\%, competitive with larger codebooks.}
\label{tab:ablation-align-ddos}
\small
\begin{tabular}{cccc}
\toprule
\multicolumn{2}{c}{\textbf{Codebook}} & \multicolumn{2}{c}{\textbf{Alignment Quality}} \\
\cmidrule(lr){1-2} \cmidrule(lr){3-4}
$K$ & $C$ & Slot AUC & Slot AUC ($\pm 1$) \\
\midrule
3 & 64 & 0.9627 & 0.9969 \\
3 & 128 & 0.9692 & 0.9976 \\
3 & 256 & 0.9700 & 0.9978 \\
3 & 512 & 0.9757 & 0.9978 \\
\bottomrule
\end{tabular}
\end{table}

\begin{table}[h!]
\centering
\caption{Alignment quality ablation on PortScan. K3-C128 achieves 98.5\%, outperforming DDoS due to binary feature clarity.}
\label{tab:ablation-align-portscan}
\small
\begin{tabular}{cccc}
\toprule
\multicolumn{2}{c}{\textbf{Codebook}} & \multicolumn{2}{c}{\textbf{Alignment Quality}} \\
\cmidrule(lr){1-2} \cmidrule(lr){3-4}
$K$ & $C$ & Slot AUC & Slot AUC ($\pm 1$) \\
\midrule
3 & 64 & 0.9824 & 0.9984 \\
3 & 128 & 0.9845 & 0.9984 \\
3 & 256 & 0.9871 & 0.9987 \\
3 & 512 & 0.9885 & 0.9988 \\
\bottomrule
\end{tabular}
\end{table}

Security tasks achieve uniformly high alignment (96.9--98.9\%), with marginal gains beyond K3-C128. PortScan outperforms DDoS due to the binary nature of PortScan features, which can be described more precisely in natural language. Tolerant accuracy exceeds 99.6\% across all security configurations, showing that alignment errors are typically off-by-one.

\subsection{Context-efficient Configuration Ablation on Security Domains}

\begin{table}[h!]
\centering
\caption{Behavioral Code configuration impact on security task performance (\%): accuracy and macro F1 for different codebook sizes.}
\label{tab:bc-ablation-security}
\small
\begin{tabular}{l|cc|cc}
\toprule
Dataset & $K$ & $C$ & Accuracy & Macro F1 \\
\midrule
\multirow{4}{*}{\textbf{DDoS}} & 3 & 64 & 99.60 & 99.59 \\
& 3 & 128 & 99.51 & 99.51 \\
& 3 & 256 & 99.56 & 99.55 \\
& 3 & 512 & 99.60 & 99.59 \\
\midrule
\multirow{4}{*}{\textbf{PortScan}} & 3 & 64 & 99.44 & 99.43 \\
& 3 & 128 & 99.20 & 99.18 \\
& 3 & 256 & 99.34 & 99.33 \\
& 3 & 512 & 99.13 & 99.11 \\
\bottomrule
\end{tabular}
\end{table}

Security tasks show robustness across K3 configurations (99.1--99.6\% F1) with minimal variation, indicating that network intrusion patterns are more forgiving of codebook size changes. We select K3-C64 for security tasks to balance model capacity with performance while maintaining high accuracy.



\end{document}